\documentclass{ecai}
\usepackage{times}
\usepackage{graphicx}
\usepackage{caption}
\usepackage{subcaption}
\usepackage{latexsym}
\usepackage{algorithm}
\usepackage[noend]{algpseudocode}
\def\BState{\State\hskip-\ALG@thistlm}

\usepackage{amsmath}
\DeclareMathOperator*{\argmax}{max}
\DeclareMathOperator*{\argmin}{min}

\DeclareMathOperator*{\argminA}{argmin}

\usepackage{mathptmx}
\usepackage{url}

\begin{document}
	
	
	\title{\LARGE Improving Solution Quality of Bounded Max-Sum Algorithm to Solve DCOPs involving Hard and Soft Constraints 
	}
	
	\author{Md. Musfiqur Rahman\institute{University of Dhaka,
			Bangladesh, Email: musfiq14shohan@gmail.com} \and Mashrur Rashik\institute{University of Dhaka,
			Bangladesh, Email: mashrur639@gmail.com} \and
		Md. Mamun-or-Rashid\institute{University of Dhaka,
			Bangladesh, Email: mamun@cse.du.ac.bd} \and
		Md. Mosaddek Khan\institute{University of Dhaka,
			Bangladesh, Email: mosaddek@du.ac.bd}}

	\maketitle
	\bibliographystyle{ecai}

	\begin{abstract}
		Bounded Max-Sum (BMS) is a message-passing algorithm that provides approximation solution to a specific form of decentralized coordination problems, namely Distributed Constrained Optimization Problems (DCOPs). In particular, BMS algorithm is able to solve problems of this type having large search space at the expense of low computational cost. Notably, the traditional DCOP formulation does not consider those constraints that must be satisfied (also known as hard constraints), rather it concentrates only on soft constraints. Hence, although the presence of both types of constraints are observed in a number of real-world applications, the BMS algorithm does not actively capitalize on the hard constraints. To address this issue, we tailor BMS in such a way that can deal with DCOPs having both type constraints. In so doing, our approach improves the solution quality of the algorithm. The empirical results exhibit a marked improvement in the quality of the solutions of\:large\:DCOPs. 
	\end{abstract}
	
	
	\vspace{-3mm}
	\section{INTRODUCTION}	
	Distributed Constrained Optimization Problems (DCOPs) are a popular framework to coordinate interactions in cooperative multi-agent systems. A number of real world problems such as distributed event scheduling \cite{maheswaran2004taking} and the distributed RLFA problem\cite{cabon1999radio} can be modelled with this framework\cite{rashik2020speeding,khan2018near,mahmud2019aed,choudhury2020particle,khan2018generic,khan2018speeding}. The constraints among the participating agents in these applications, and many other besides, can be both hard and soft. In any case, the algorithms that have been proposed to solve DCOPs can be broadly classified into exact and non-exact algorithms. The former (e.g.\cite{modi2005adopt,petcu2005scalable}) always finds a globally optimal solution. In contrast, the latter algorithms (e.g. \cite{farinelli2008decentralised,zhang2005distributed}) trade solution quality at the expense of reduced computation and communication costs. 
	
	Among the non-exact approaches, Generalized Distributive Law based algorithms, such as Max-Sum \cite{farinelli2008decentralised} and Bounded Max-Sum (BMS) \cite{rogers2011bounded}, have received particular attention. Specifically, Bounded Max-Sum is extremely attractive variant of Max-Sum which produces good approximate solution for DCOP problems with cycles. However, BMS does not actively consider such constraints that are hard although there is a number of real-life applications containing hard constraints. We particularly observe that 
    the presence of hard constraints can be utilized to further improve BMS's solution quality by removing inconsistent values from agents' domain and thus reduce the upper bound of the global solution. It is worth noting that due to hard constraints the traditional BMS algorithm exhibits a situation where each agent may own a set of allowable assignments in place of a specific assignment. Each combination of the agent's assignments will experience the same profit for the tree structured graphical representation that is a acyclic graph (e.g. Factor Graphs or Junction Tree) of a given DCOP, but produce different profit for that cyclic DCOP. To the best of our knowledge, there exists no method for choosing the best one. In this paper, we propose a novel approach which aims at enforcing consistency and selecting the most preferable combination of agent's assignments.
	\vspace{-5mm}
	\section{PROBLEM FORMULATION}
	
	\par A DCOP model can be formally expressed as a tuple $\langle$\textbf{A}, \textbf{X}, \textbf{D}, \textbf{F}, $\alpha\rangle$ where \textbf{A} = \{$a_1,a_2,....,a_n$\} is a set of agents, \textbf{X} = \{$x_1,x_2,....,x_n$\} is a set of variables, \textbf{D} = \{$d_1,d_2,....,d_n$\} is a set of domains for the variables in \textbf{X}. \textbf{F} = \{$f_1,f_2,....,f_m$\} is a set of constraint functions.
	$f_j( \mathbf{x_i})$ denotes value for each possible combination of the variables of $\mathbf{x_i} \in X$. The dependencies between the functions and variables can be graphically represented by factor graph $\textbf{FG}$. Finally, the mapping of variable node to agent is represented by $\alpha$ : \textbf{X} $\rightarrow$ \textbf{A} where one variable will be assigned to one agent.
	\par Within this model, the main objective of DCOP algorithms such as BMS, is to find the assignment of each variable, $\mathbf{\tilde{x}}$ and approximate solution $\tilde{V}$ by maximizing the sum of all functions that is \begin{math}\tilde{V}= \sum_{j}^{m} f_{j}(\tilde{x}_{i}) \end{math}. After removing appropriate dependencies from $\textbf{FG}$ according to the phases of BMS, an acyclic graph $\mathbf{\overline{FG}}$ is formed. Additionally, the maximum impact $B$ is calculated which is used for computing the upper bound on the value of the unknown optimal solution as $\tilde{V}^m+B$. Here $\tilde{V}^m$ is the solution found by executing BMS on the corresponding acyclic graph.
	%
	%
	Now, the first objective of our approach is to update the domain of each variables so that the maximum impact is minimized as in (Equation~\ref{CE}). Here,  $\tilde{d_{i}}$ = inconsistent domain values of variable $x_i$.
		\vspace{-4mm}
		\begin{equation}
	\begin{aligned}
	D=\argminA_{d_1,...,d_n}  B \\
	s.t. \forall d_i \in D , d_i= d_i \char`\\ \tilde{d_{i}} \\
	\end{aligned}
	\label{CE}
	\end{equation}
			\vspace{-0.5mm}
Due to the presence of hard constraints, the BMS algorithm experiences tie variable assignment(s) after executing Max-Sum on its acyclic graph. Our second objective is to select appropriate variable assignment so that it provides the most preferable solution for the constraint graph (Equation~\ref{tie}). Here, $T_i \subseteq  D_i$ is set of tie assignments for variable $x_i$. 
		\vspace{-5mm}
\begin{equation}
	\begin{aligned}
	\tilde{\textbf{x}}=\argminA_{x_1,...,x_n} \sum_{j}^{m} f_j(\mathbf{x_i}) \\
	s.t. (x_1,x_2,...,x_n) \in (T_1\times T_2\times,...,\times T_n) \and 
	\end{aligned}
	\label{tie}
	\end{equation}
	\begin{figure*}[ht]
		\centering
		\begin{subfigure}[H]{.3\textwidth}
			\centering
			\includegraphics[scale=0.5]{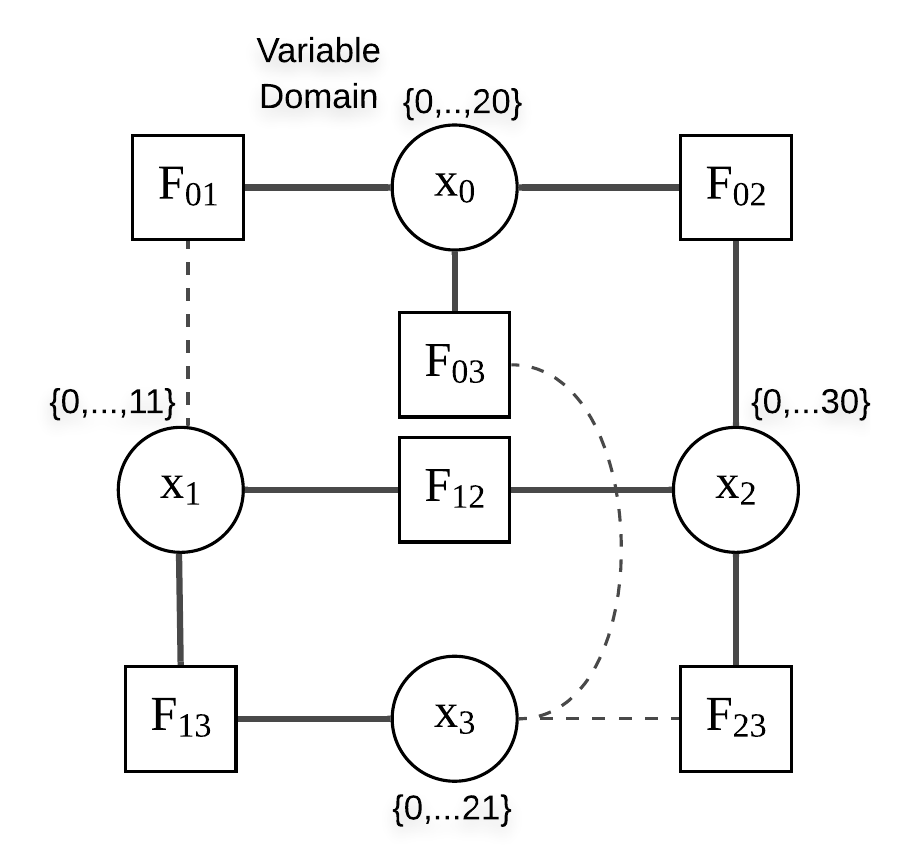}
			\caption{Arc consistency is enforced and domain is pruned. Maximum impact reduced from 1473 to 1457. A spanning tree (acyclic graph) is created from the cyclic factor graph.
				 } \label{graph1}
		\end{subfigure}
		\hfill
		\begin{subfigure}[H]{.3\textwidth}
			\centering
			\includegraphics[scale=0.5]{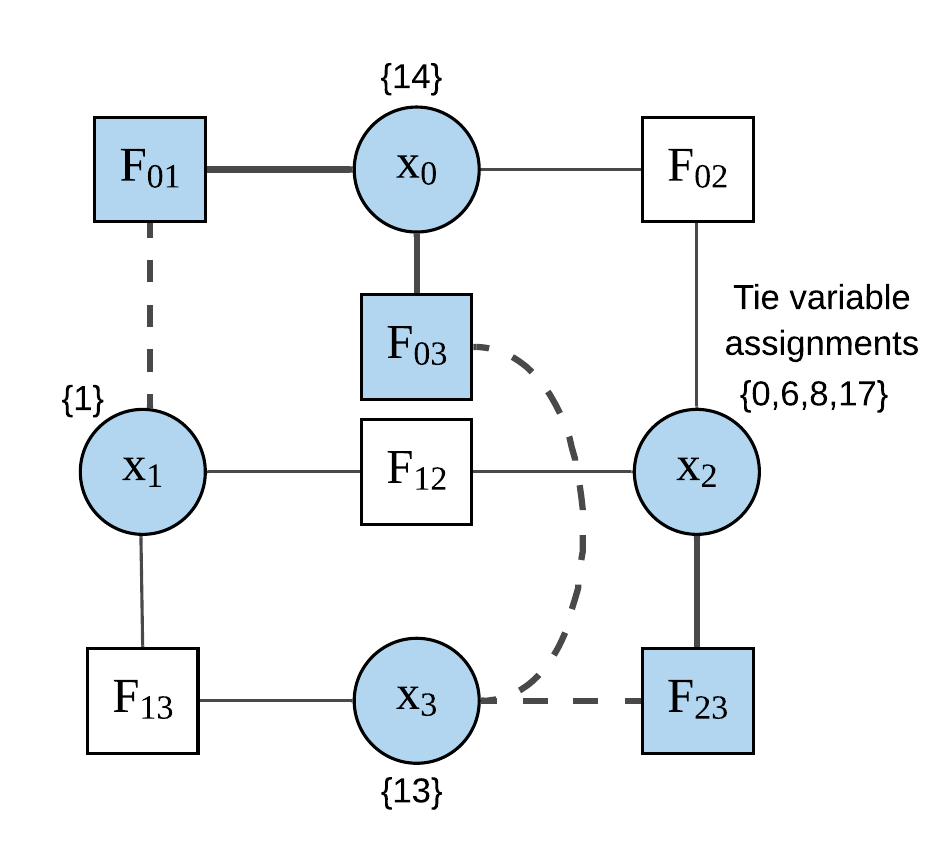}
			\caption{Tie assignment is found for each variable. Next,new DCOP problem is created including the blue colored nodes and BMS is executed on these nodes.	
				 } \label{graph2}
		\end{subfigure}
		\hfill
		\begin{subfigure}[H]{.3\textwidth}
			\centering
			\includegraphics[scale=0.5]{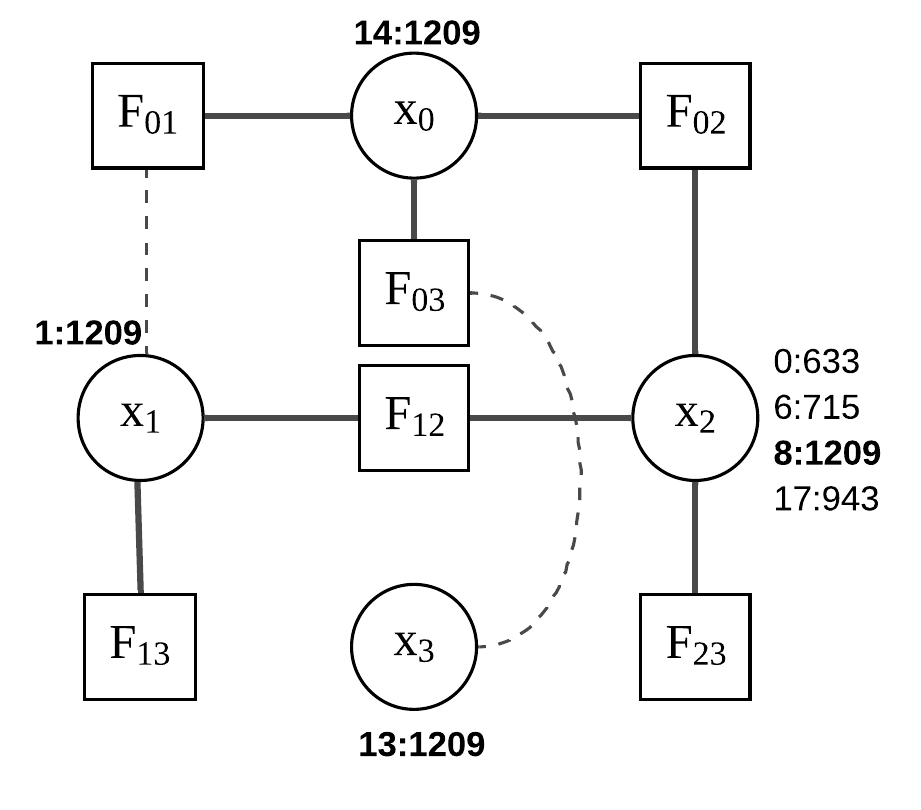}
			\caption{
				Priority for each tie assignment is found. Using this information, BMS is executed on the main acyclic graph.
				 } \label{graph3}
		\end{subfigure}
		\vspace{-3mm}
		\caption{Worked example of the HBMS algorithm. Here the square nodes and round nodes represent function nodes and variable nodes, respectively. The thick edges of the graph are our main points of interest. The third phase is executed on the sub-graph phase consists of blue color nodes.
			processes } \label{hbms_simulation}
	\end{figure*}
	
	\vspace{-5mm}
	\section{HARD CONSISTENCY ENFORCED BOUNDED Max-Sum (HBMS)}
	With the motivation of utilizing hard constraints, our objective is to 
	decrease the upper bound on the optimal solution and also increase the solution quality of BMS. Our first contribution in the upper bound is obtained by changing the maximum impact $B$. In  our \textbf{first phase}, consistency enforcement, we update the variable domains (Equation~\ref{CE}) by enforcing arc-consistency on the constraint graph. This step also speeds up the execution time of the algorithm. For example, in Figure~\ref{graph1}, $F_{01}$\footnote{In this paper, we have considered all constraints $f_i$ as binary and mentioned as $F_{x_i,x_j}$ for illustration where $x_i,x_j$ are dependent on $f_i$} calculates its maximum impact $B_{01}$ using Equation~\ref{upperbound}. According to this equation, $x_1$ selects its value $x_{1h}$ for maximizing and $x_{1l}$ for minimizing the $F_{01}$ function. After the consistency enforcement phase, each variable's domain gets pruned and ($x_{1h},x_{1l}$) pair changes to ($x_{1h}^\prime,x_{1l}^\prime$) by reducing $B$. This is true for $F_{03}$ and $F_{23}$ in the same way.
			\vspace{-3mm}
	\begin{equation}
	B_{01}=\argmax_{x_0} \Bigg[ \argmax_{x_{1}}F_{01}(x_0,x_1)  -\argmin_{x_{1}}F_{01}(x_0,x_1) \Bigg]
	\label{upperbound}
	\end{equation}  
	
	In the \textbf{second phase}, we generate a spanning tree (i.e. acyclic graph) from the factor graph by removing the most suitable dependencies. We do this following the same way as the BMS algorithm. Then run Max-Sum algorithm on the acyclic factor graph. If the hard constraints are satisfied, each of them will contribute the same in profit maximization. This phenomenon increases the possibility for each variable having multiple assignment ($T_i$ is the set of assignment allowed for variable $x_i$) with same profit for the acyclic graph.
	For example, in Figure~\ref{graph2}, after executing Max-Sum, $x_2$ is assigned with multiple values that is $\{0,6,8,17\}$.
	However, we need to chose the variable assignments in such a way that the profit for the main constraint graph is maximized (Equation~\ref{tie}). For this purpose,	we utilize the removed dependencies and the set of variables $\mathbf{x_i^c}$ that are dependent on function $F_i$  but are not a part of the acyclic graph $\mathbf{\overline{FG}}$. In the \textbf{third phase}, we model a smaller DCOP problem as $\langle \mathbf{A}', \mathbf{X}', \mathbf{D}', \mathbf{F}', \alpha\rangle$ such that,  \textbf{F} = \{$f_1,f_2,....,f_k$\} where $\forall F_i, \mathbf{x_i^c} \neq \O$ and select a set of agents $\mathbf{A}'$ and variables $\mathbf{X}'$ dependent on the function accordingly. Finally, for each $d_i \in \mathbf{D}$, $d_i$ will be equals to $T_i$. According to Figure~\ref{graph2}, we create a new DCOP and represents it as a factor graph but this time it includes the dependencies that are removed in the previous phase. It consists of $F_{01}, F_{03}, F_{23}$ and their corresponding variables ($x_0,x_1$) , ($x_0,x_3$) and ($x_2,x_3$). At this phase, we execute BMS on this smaller graph and eventually get information about the priority of the tie assignments of each variables that we found in second phase. In Figure~\ref{graph3}, we can see the profit for each assignment received from the third phase (e.g. for $x_2, (0:633),(6:715),(8:1209),(17:943)$). Finally, In the \textbf{fourth phase}, we use this information to execute Max-Sum on $\mathbf{\overline{FG}}$ again. This step will finally select the preferable variable assignment which improves solution quality. For instance, the variable assignment is $(x_0:14),(x_1:1),(x_2:8) \and (x_3:13)$. The complexity of HBMS is twice of the BMS algorithm since we execute this algorithm two times. The computation cost for the smaller graph in the second phase is negligible. Finally, the complexity of the arc consistency enforcement phase is $ed^3$ where $e$ is the number of edges of the constraint graph and $d$ is the average domain size.

\section{EMPIRICAL EVALUATION}
\begin{figure}[t]
		\centering
		\vspace{-5mm}
		\includegraphics[scale=0.4]{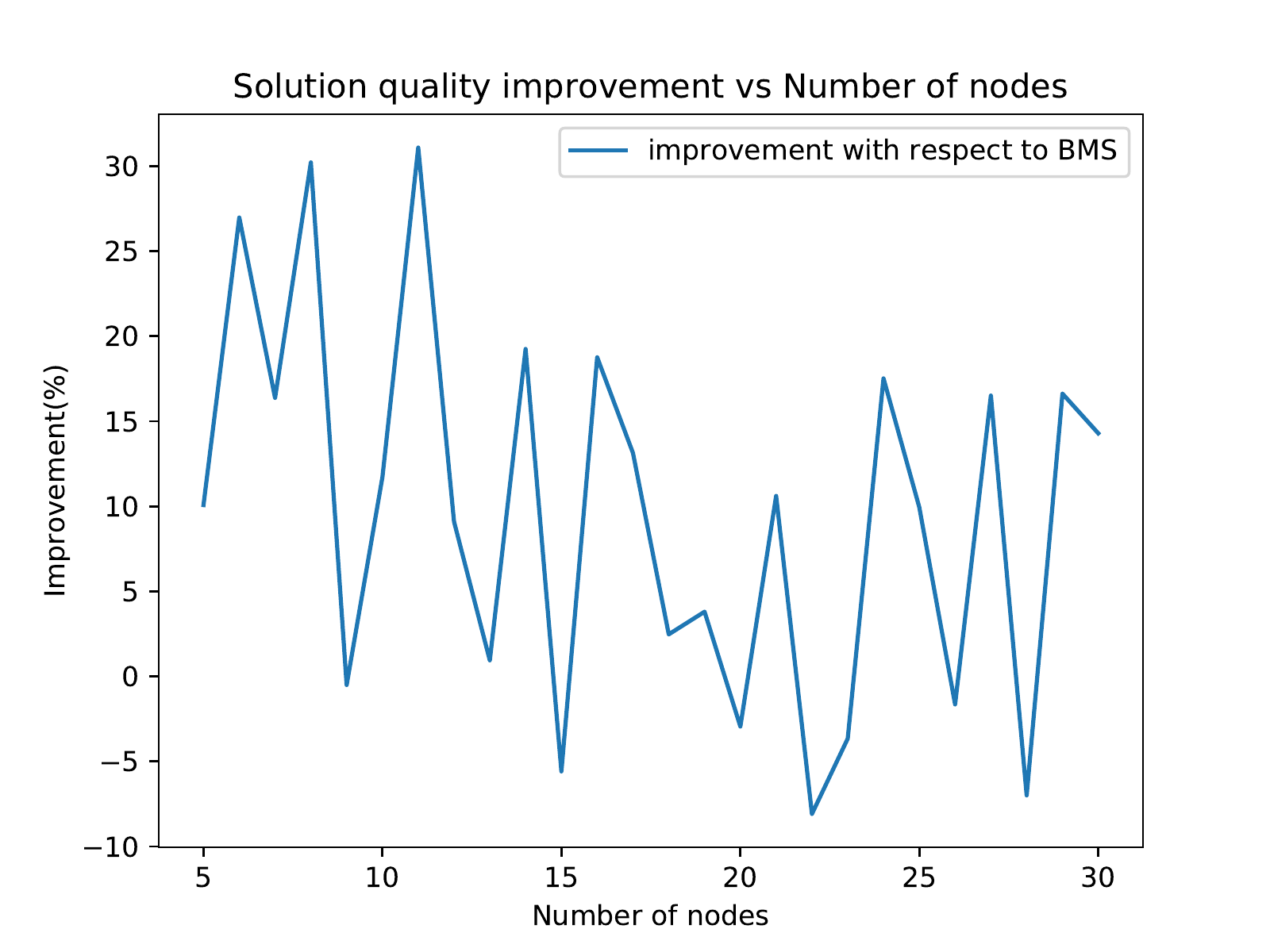}
		\vspace{-3.4mm}
		\caption{Empirical result for constraint graphs varying number of nodes from 4 to30. Improvement is calculated in percentage with respect to Bounded Max-Sum.}
		\label{performance}
	\end{figure}

	In this section, we empirically evaluate the improvement in solution quality of HBMS in comparison to the Bounded Max-Sum algorithm. To benchmark the result, we run experiment on random constraint graphs. We vary the number of nodes from 5 to 30 in Figure~\ref{performance}, set the variable domain in [0,..,40], 30\% hard constraints along with soft constraints and functions' utility values from 0 to 500. We observe improvement in solution quality around 5-30\% on average. However, we experience negative results in some instances. In the future, we would like to explore that area for observing the reasons behind this situation.

		
    \vspace{-1mm}	
	\section{CONCLUSIONS AND FUTURE WORK}
    The major finding of this paper is that by taking advantage of the hard constraints, we can significantly improve the solution quality of the Bounded Max-Sum algorithm. Another notable contribution is in the reduction of the upper bound. Our empirical evidence presents that it is possible to improve the solution around 10-30\% than BMS. In our future work, we would to like to observe the impact of different forms of consistency enforcement, and fix the negative results observed in the evaluation. The final research direction includes extending the potential application domain.

	\vspace{-3mm}
	\bibliography{bibliography}

	\end{document}